\pdfoutput=1

\documentclass[11pt]{article}

\usepackage{emnlp2021}

\usepackage{times}
\usepackage{latexsym}

\usepackage[T1]{fontenc}

\usepackage[utf8]{inputenc}

\usepackage{microtype}

\usepackage{bm}
\usepackage{amsmath}
\usepackage{booktabs}
\usepackage{algorithm}  
\usepackage{algorithmicx}
\usepackage{appendix}
\usepackage{algpseudocode}  
\usepackage{amsmath} 
\usepackage{url}

\usepackage{listings}
\usepackage{color}
\usepackage{graphicx, subfig}

\definecolor{dkgreen}{rgb}{0,0.6,0}
\definecolor{gray}{rgb}{0.5,0.5,0.5}
\definecolor{mauve}{rgb}{0.58,0,0.82}

\usepackage{microtype}

\title{Constructing Contrastive Samples via Summarization for Text Classification with Limited Annotations}

\author{Yangkai Du\textsuperscript{1}, Tengfei Ma\textsuperscript{2}, Lingfei Wu\textsuperscript{3}, Fangli Xu\textsuperscript{4}, Xuhong Zhang\textsuperscript{1} \\\bf{Bo Long}\textsuperscript{3}, \bf{Shouling Ji}\textsuperscript{1} \\
        \textsuperscript{1}Zhejiang University; \textsuperscript{2}IBM Research; 
        \textsuperscript{3}JD.COM; \textsuperscript{4}Squirrel AI Learning \\
        \texttt{\{{yangkaidu,zhangxuhong,sji}\}@zju.edu.cn} \\
        \texttt{tengfei.ma1@ibm.com, \{lingfei.wu,bo.long\}@jd.com, lili@yixue.us}
        }

\begin{document}
\maketitle
\begin{abstract}
 Contrastive Learning has emerged as a powerful representation learning method and facilitates various downstream tasks especially when supervised data is limited. How to construct efficient contrastive samples through data augmentation is key to its success. Unlike vision tasks, the data augmentation method for contrastive learning has not been investigated sufficiently in language tasks. In this paper, we propose a novel approach to construct contrastive samples for language tasks using text summarization. We use these samples for supervised contrastive learning to gain better text representations which greatly benefit text classification tasks with limited annotations. To further improve the method, we mix up samples from different classes and add an extra regularization, named Mixsum, in addition to the cross-entropy-loss.
Experiments on real-world text classification datasets (Amazon-5, Yelp-5, AG News, and IMDb) demonstrate the effectiveness of the proposed contrastive learning framework with summarization-based data augmentation and Mixsum regularization.
\end{abstract}

\section{Introduction}
Learning a good representation has been an essential problem in the deep learning era. Especially, in the area of natural language processing, the language model pre-training techniques, such as BERT \cite{devlin-etal-2019-bert}, have been overwhelming in a wide range of tasks by learning contextualized representations. However, the success of these pre-trained models hinge largely on plenty of labeled data for fine-tuning. With limited labels on the target task, fine-tuning BERT has been shown unstable\cite{zhang2020revisiting}. In practice, it is costly to gather labeled data for a new task, and lack of training data is still a big challenge in many real-world problems.

Recently, contrastive learning methods have become popular self-supervised learning tools and gained big progress in few-shot learning due to its better discriminative ability \cite{gidaris2019boosting,su2020does}. Various contrastive learning methods have been developed and lead to state-of-the-art performance in many computer vision tasks. They are also extended to the fully supervised setting by leveraging label information to make further improvement. In natural language processing, contrastive learning has not been fully investigated but it is attracting more and more attentions.

A contrastive learning method generally consists of two components: finding positive samples and negative samples for each anchor sample; and building up an effective objective function to discriminate them. In many contrastive learning frameworks, how to efficiently find the contrastive samples has been the key to their success. For example, in MoCo\cite{he_momentum_2020}, the contrastive pairs are constructed by matching an encoded query with a dynamic dictionary; in SimCLR\cite{pmlr-v119-chen20j}, the contrastive pairs are created by applying two different data augmentation operators, and it was shown that composition of data augmentation operations is crucial for learning good representations. In supervised contrastive learning, essentially the positive sample space has been augmented. Instead of only using the anchor sample and its own transformation, all samples in the same class can be further regarded as positive pairs. 

In this paper, we focus on using contrastive learning to assist the text classification tasks with limited labels. Considering the specialty of the text classification task, we propose two novel strategies to further enhance the performance of supervised contrastive learning. We assume that a good summarization system can keep the most critical information of original texts and the generated summary tends to belong to the same category as the original text. Thus we utilize text summarization as a data augmentation method to create more positive and negative samples for supervised contrastive learning. Furthermore, we propose Mixsum, an idea similar to the methodology of mix-up\cite{DBLP:conf/iclr/ZhangCDL18}, which combines texts from different categories and creates new summary samples to further augment the data for contrastive learning. We adapt the supervised contrastive loss to the Mixsum setting, and show that it brings great benefit for text classification when training data is extremely scarce. 

Our main contributions are listed as below:
\begin{itemize}
    \item We propose a new contrastive learning framework for text representation learning and mitigate the label deficiency problem for text classification.
    \item We employ text summarization, a new data augmentation method, to construct positive and negative sample pairs for contrastive learning. 
    \item We improve the supervised contrastive learning method by mixing up the samples in different categories. Combining with the summarization based data augmentation method, our model shows superior performance on three real-world datasets.
\end{itemize}


\section{Background and Related Works}
\label{sec:background}
\subsection{Contrastive Learning}
The main idea of contrastive learning is minimizing the vector distance between anchor examples and positive examples while maximizing the vector distance between anchor examples and negative examples.

Self-supervised contrastive Learning has been demonstrated effective on many computer vision tasks \cite{he_momentum_2020,pmlr-v119-chen20j}. In a self-supervised contrastive learning framework, anchor samples are the original data samples, positive samples are the augmented anchor sample, and negative samples are generally set to all other samples in the mini-batch.
\begin{equation}
    \label{eq:self-supervised-obj}
    L_{self} = \sum _{i=1}^{N} -log \frac{exp(f(x_i)\cdot f(x_{2i})/ \tau)}{\sum _{k=1}^{2N} 1_{i\neq k} exp(f(x_i)\cdot f(x_{k})/ \tau) } 
\end{equation}
Equation \ref{eq:self-supervised-obj} is the self-supervised contrastive learning objective for the popular SimCLR framework \cite{pmlr-v119-chen20j}.
For each mini-batch with N anchor samples, we can get another N positive samples by data augmentation, concatenate them to form a new batch. Then for each anchor examples index, $i$ in the range $\{1,2,...,N\}$, the index for the corresponding positive sample is $2i$, and all other $2N-2$ samples in the batch are negative samples.
$f(\cdot)$ is a representation model mapping the input samples to a normalized dense vector in $R^d$, and $\tau$ is the temperature parameter. 
Contrastive learning on NLP tasks also arises much research intensity recently.~\citet{fang_cert_2020} propose to learn sentence-level representations by fine-tuning BERT\cite{devlin-etal-2019-bert} with back-translation based data augmentation and self-supervised contrastive learning objective function. ~\citet{klein-nabi-2020-contrastive} propose to use contrastive learning for commonsense reasoning, and the proposed method alleviates the current limitation of supervised commonsense reasoning.  \citet{NEURIPS2020_d89a66c7} explore the general supervised contrastive learning loss and show the effectiveness of supervised contrastive learning. \citet{DBLP:journals/corr/abs-2011-01403} introduced the supervised contrastive loss to the original cross-entropy loss for fine-tuning pre-trained transformers like Roberta\cite{DBLP:journals/corr/abs-1907-11692} and BERT\cite{devlin-etal-2019-bert}, which is highly related to our work. Our approach is different from these previous works in that we utilize a new data augmentation, i.e. summarization, for supervised contrastive learning. Our Mixsum method is also never explored by those methods.

\subsection{Beyond Empirical Risk Minimization}
The general theme of supervised learning is minimizing the empirical risk of datasets by defining a loss function $l$, which describes the difference between the model prediction $f(x)$ and target label $y$. The expected risk of the datasets can be described in Equation \ref{eq:erm-int}.
\begin{equation}
    \label{eq:erm-int}
    R(f) = \int l(f(x),y) dP(x,y)
\end{equation}
P(x,y) is the distribution of the dataset, which is unknown but can be approximated by empirical distribution. Then we can now approximate the expected risk by empirical risk in Equation \ref{eq:emperical-risk}.
\begin{equation}
    \label{eq:emperical-risk}
    R_e(f) = \frac{1}{n} \sum _{i=1}^{N} l(f(x_i),y_i)
\end{equation}
Minimizing the empirical risk in Equation \ref{eq:emperical-risk} is called Empirical Risk Minimization(ERM) \cite{788640}. ERM will lead the model to memorize the training samples and fail for data out of training samples. Motivated by the limitation of ERM, \citet{DBLP:conf/iclr/ZhangCDL18} propose a generic vicinal distribution, called mixup:
\begin{equation}
    \label{eq:mixup}
    \begin{aligned}
    \tilde{x} = &\lambda x_i + (1-\lambda) x_j \\
    \tilde{y} = &\lambda y_i + (1-\lambda) y_j
    \end{aligned}
\end{equation}
\citet{DBLP:conf/iclr/ZhangCDL18} use this new vicinal distribution described in Equation \ref{eq:mixup} to approximate the expected risk, and minimizing the empirical vicinal risk\cite{NIPS2000_ba9a56ce} in Equation \ref{eq:vrm}.
\begin{equation}
    \label{eq:vrm}
    R_v(f) = \frac{1}{n} \sum _{i=1}^{N}l(f(\tilde{x_i}), \tilde{y_i})
\end{equation}

The proposed vicinal distribution--mixup, can be viewed as a form of data augmentation that leads the model to behave in between the training samples and soften the labels. Experiments demonstrate that mixup can improve the robustness of the trained model and avoid undesirable oscillations when predicting unseen samples\cite{DBLP:conf/iclr/ZhangCDL18}.

Besides, Kim et.al \cite{kim_mixco_2020} proposed MixCo, which create a vicinal distribution for self-supervised contrastive learning based on the idea of mixup\cite{DBLP:conf/iclr/ZhangCDL18}, they demonstrate the effectiveness of vicinal distribution minimization for self-supervised contrastive learning loss over image classification tasks. Inspired by mixup and MixCo, we propose a novel vicinal distribution, i.e. Mixsum, for supervised contrastive learning.

\section{Methods}
\label{sec:method}

\subsection{Problem Definition}
The task we want to solve is text classification with limited annotations. In the text classification task, the input data is usually a sentence, a paragraph or a document. Assume we have a small number of training samples with labels $D_{train}$ and a large amount of unlabeled data $D_{test}$. For each text sample $x \in D_{train}$, it has a label $y$ which is from $L$ classes. And we want to predict the labels of all samples in the test data.

\begin {figure*}[t]
	\centering
	\includegraphics[width=15cm]{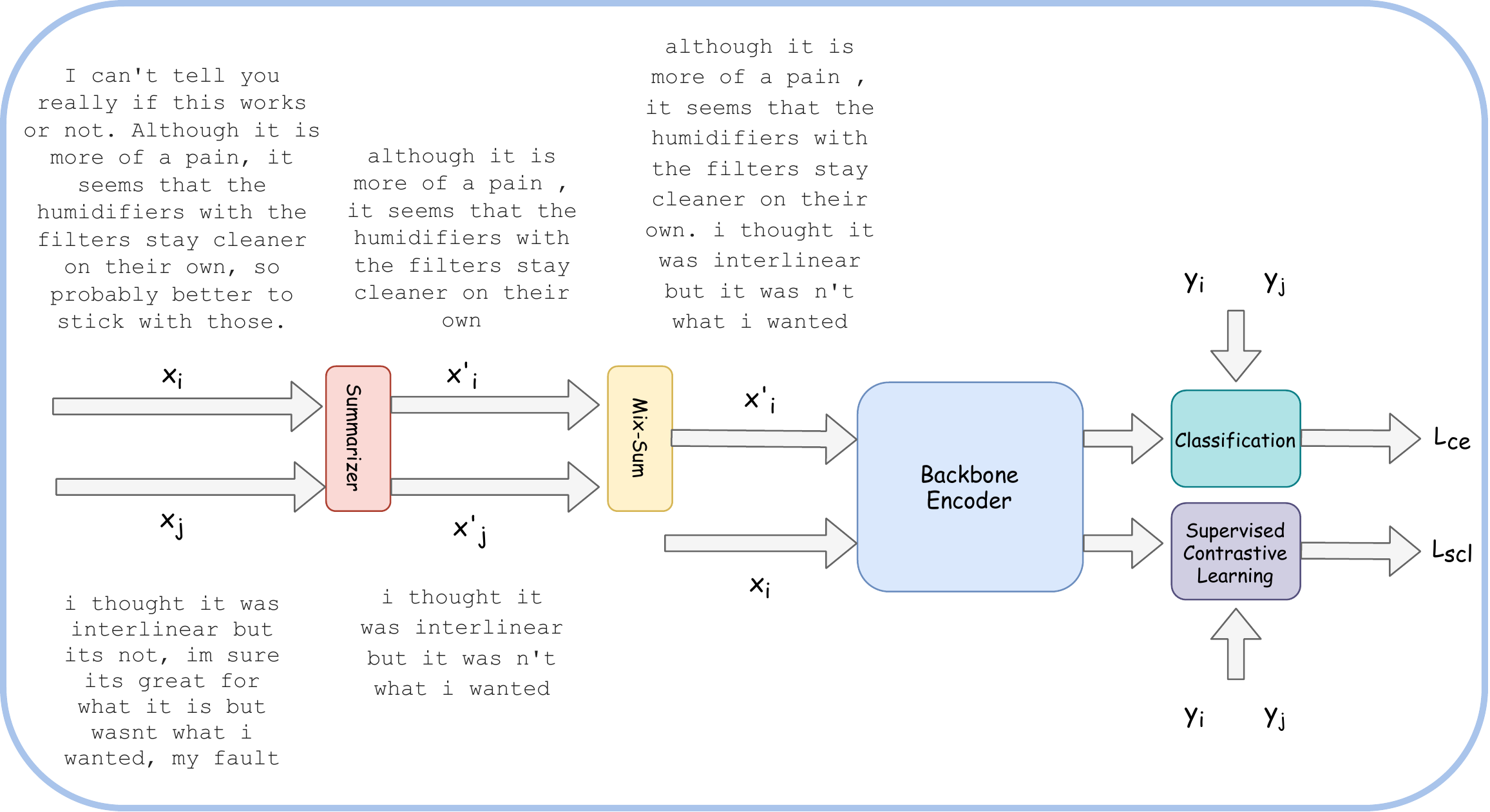}
	\caption{Illustration of using summaries as contrastive samples for text classification. $x_{i}$ is the original text, $x'_i$ is the summary of $x_i$, $y_i$ is the target label for $x_i$. Randomly select another sample $x_j$, concatenate the summary of $x_j$--$x'_j$ with $x'_i$, and use it as the contrastive sample of $x_i$}
	\label{fig:illustration}
	\vspace{-0.5cm}
\end{figure*}

\subsection{Text Summarization}
We propose to use text summarization as the data augmentation strategy for constructing positive and negative samples in supervised contrastive learning when the number of annotated training samples is limited. Intuitively, the summarization process can filter out unnecessary and redundant information in the text and extract the most representative semantics. The summary owns the same label as its source text.

We use PreSumm \cite{liu-lapata-2019-text} for automatic text summarization. PreSumm utilizes BERT as a general framework for both extractive and abstractive summarization, both of them can achieve great summarization quality even without text-summary pairs for finetuning. For each input text $x$ we can get its summary $x'$ by feeding the input text $x_i$ to PreSumm model \ref{summarization},where $i$ is the index in Minibatch.
\begin{equation}
    \label{summarization}
    x_i' = PreSumm(x_i)
\end{equation}

We use the abstractive summarization model trained by \cite{liu-lapata-2019-text} without any text-summary pairs for fine-tuning. Compared to extractive summarization, which can only generate summaries by extracting key sentences from original paragraphs, abstractive summarization can generate information-rich, coherent and less-redundant summary compared to extractive summary and do not have the limitation that summary is only from the original text. 

Assuming the generated summaries belong to the same class as their original source texts, we can add them to the training samples. 

\subsection{Supervised Contrastive Learning}
\label{sec:methods-sup}
Although fine-tuning pretrained model using cross-entropy is commonly used for text classification, and it achieves state-of-art results on many text classification tasks\cite{NEURIPS2019_dc6a7e65}. However, this approach still can not achieve optimal performance in few-shot setting, where training data is limited. In order to alleviate this limitation, we propose to add a supervised contrastive learning objective \cite{DBLP:journals/corr/abs-2011-01403} and using text summaries as contrastive samples to train a more robust text classifier under the limited annotation setting.

The main idea of supervised contrastive learning is minimizing the intra-class representation distance while maximizing the inter-class representation distance. It would be easier for the classifier to learn a good decision boundary by applying supervised contrastive learning. This process can be achieved by minimizing Equation \ref{sup-con}.

For each batch with $N$ input texts and $N$ labels, we first apply summarization to get the augmented $N$ text summaries; then, we get $2N$ samples in a batch. For each anchor sample $x_i$, we want to minimize the vector distance between $x_i$ and positive samples $x_j$, whose labels $y_i$ and $y_j$ belong to the same class.
\begin{equation}
\begin{aligned}
    \label{sup-con}
     L_{sup}&(X,Y) = - \frac{1}{2N} \sum_{i=1}^{2N} \frac{1}{N_{y_{i}} - 1} \sum _{j=1}^{2N} \\
     & 1_{i\neq j}1_{y_i = y_j} \frac{exp(g(x_i) \cdot g(x_j)/\tau)}{\sum _{k=1}^{2N}1_{k\neq i} exp(g(x_i) \cdot g(x_k)/\tau)}
\end{aligned}
\end{equation}
Where $N$ is the mini-batch size, and $2N$ is the size of the augmented batch after applying summarization. $N_{y_i}$ is number of samples which have same labels as $y_i$. Labels for the summary is the same as the original text. $X$ and $Y$ are the batches of augmented training samples and target labels. $g(\cdot)$ is $l_2$ normalized representation of input text in $\mathbf{R}^n$, where $n$ is the dimension of text feature used for supervised contrastive learning. The similarity measure of $g(\cdot)$ is cosine similarity with temperature parameter $\tau$. The cosine similarity of $g(x_i)$ and $g(x_j)$ should be maximized when $x_i$ and $x_j$ come from the same class; otherwise it should be minimized.

Since contrastive learning can gain better performance when an MLP head is used \cite{he_momentum_2020}, we also apply an MLP head upon the base text encoder $\Phi(\cdot)$. The text encoder $\Phi(\cdot)$ can be any pretrained text encoder which maps a text to a dense vector in $\mathbf{R}^d$, eg. BERT\cite{devlin-etal-2019-bert}, XLNet\cite{NEURIPS2019_dc6a7e65}, Roberta\cite{DBLP:journals/corr/abs-1907-11692}, LSTMs and CNNs\cite{zhang_character-level_2015}. $d$ is the feature dimension of the text encoder. The entire text encoding process is expressed in Equation \ref{encoding} and \ref{norm}.
\begin{equation}
    \label{encoding}
    G(x) = MLP(\Phi(x))
\end{equation}
\begin{equation}
    \label{norm}
    g(x) = \frac{G(x)}{||G(x)||}
\end{equation}

Combining the cross-entropy loss in Equation \ref{eq:ce} with a trade-off parameter $\lambda$, we can get the final loss function in Equation \ref{eq:total-loss}. $\lambda$ is a hyperparameter to control the relative importance of cross-entropy loss and supervised contrastive loss.
\begin{equation}
    \label{eq:total-loss}
    L(X,Y) = \lambda \cdot L_{ce}(X,Y) + (1-\lambda) \cdot L_{sup}(X,Y)
\end{equation}

\begin{equation}
    \label{eq:ce}
    \begin{aligned}
    L_{ce}& (X,Y) =  -\frac{1}{2N} \sum _{i=1}^{2N} \\
    & \mathbf{y_i} log(\mathbf{p}(x_i)) + (1 - \mathbf{y_i})log(1-\mathbf{p}(x_i))
\end{aligned}
\end{equation}
\begin{equation}
    \mathbf{p}(x_i) = \text{Softmax}(\mathbf{W}\cdot\Phi(x_i) + \mathbf{b})
\end{equation}
where $\mathbf{y_i}$ is the label of training sample $x_i$ in one-hot representation. $\mathbf{p}(x_i)$ is the predicted probability distribution generated by the text classification model. $\Phi(\cdot)$ is the backbone text encoder, which is exactly the same as the text encoder used in the supervised contrastive learning stage and the model weights also shared in supervised contrastive learning stage. $\mathbf{W}$ is a fully connected classification projection matrix in $\mathbf{R}^{C\times d}$, which map the text feature in $\mathbf{R}^d$ to score vector of output classes in $\mathbf{R}^C$. $\mathbf{b}$ is the bias of the classification head in $\mathbf{R}^C$. $C$ is the number of different classes across the training samples.

\subsection{Mixsum}
\label{sec:methods-mixsum}
We propose another novel method, i.e. Mixsum, by combining the idea of mix-up \cite{DBLP:conf/iclr/ZhangCDL18} and using summarization to construct contrastive samples--to achieve better text classification performance under the limited annotation setting. Basically, the main idea is that summaries of concatenated texts from different classes contain the feature of both classes, then the newly generated summary can serve as the regularization for cross-entropy loss and supervised contrastive learning objective, which can lead the model to behave in between the training samples and soften the labels. 

Similar to mixup \cite{DBLP:conf/iclr/ZhangCDL18}, which use a convex combination of the input image to create the vicinal distribution, we propose to combine the summaries of texts from two different classes and use the conjunct summary as the augmentation.

There are also other methods for mixing the texts from two different classes, such as linear interpolation of sentence-level features\cite{guo2019augmenting,sun-etal-2020-mixup} and word-level features\cite{guo2019augmenting}. Those methods are also applicable under our setting. 
In the summarization context, concatenating two documents with the same weight is the simplest and most intuitive way to keep our model neat and practical. Consequently, we choose this method for mixing up the texts and the $\lambda$ for mixing the vicinal label in Equation \ref{eq:mixup} is also fixed at 0.5.

\begin{equation}
    \hat{x_i'} = x_i' | x_j' 
\end{equation}
\begin{equation}
    \label{eq:mix-label}
    \hat{y_i} = 0.5 \cdot y_i + 0.5 \cdot y_j
\end{equation}
Where $x_i'$ is the summary of the original text $x_i$ in a batch, then randomly pick another summary $x_j'$ in the batch and conjunct them together to form a mix-up summary $\hat{x_i'}$. This process can be visualized in Figure \ref{fig:illustration}. The new generated label $\hat{y_i}$ follows the mix-up method introduced in \cite{DBLP:conf/iclr/ZhangCDL18}. 

Same as the contrastive samples augmentation strategy mentioned in Section \ref{sec:methods-sup}, we concatenate the original $N$ input texts with the mix-up summaries to form a new Minibatch with $2N$ samples.
Then we can formulate the new cross-entropy loss and supervised contrastive loss under Mixsum setting in Equation \ref{eq:mix-ce} and \ref{eq:mix-sup-simplified}.
\begin{equation}
    \label{eq:mix-ce}
    \begin{aligned}
     &L_{ce}^{mix}(X,Y) = -\frac{1}{2N}\cdot \\ & (\sum_{i=1}^{N} \bm{y_i} log(\bm{p}(x_i)) + (1 - \bm{y_i})log(1-\bm{p}(x_i)) \\ & \sum_{i=1}^{N} \bm{\hat{y_i}} log(\bm{p}(\hat{x_i'})) + (1 - \bm{\hat{y_i}})log(1-\bm{p}(\hat{x_i'})))
    \end{aligned}
\end{equation}
The first $N$ samples in the Minibatch $X$ are original texts, and the loss of those N samples remains the same as the cross-entropy loss. The later N samples in the Minibatch are mix-up summary.

Taking the Equation \ref{eq:mix-label} to Equation \ref{eq:mix-ce}, we can further get the compact form for the cross entropy loss under Mixsum setting in Equation \ref{eq:mix-ce-compact} and \ref{eq:mix-ce-simplified}.
\begin{equation}
\label{eq:mix-ce-compact}
    \begin{aligned}
    &L_{ce}^{mix}(X,Y) = -0.5 \cdot \frac{1}{2N} 
    \\ & (\sum_{i=1}^{N} \bm{y_i} log(\bm{p}(x_i)) + (1 - \bm{y_i})log(1-\bm{p}(x_i)) +
    \\ & \sum_{i=1}^{N} \bm{y_i} log(\bm{p}(\hat{x_i'})) + (1 - \bm{y_i})log(1-\bm{p}(\hat{x_i'})) +
    \\ & \sum_{i=1}^{N} \bm{y_i} log(\bm{p}(x_i)) + (1 - \bm{y_i})log(1-\bm{p}(x_i)) + 
    \\ & \sum_{i=1}^{N} \bm{y_j} log(\bm{p}(\hat{x_i'})) + (1 - \bm{y_j})log(1-\bm{p}(\hat{x_i'})))
    \end{aligned}
\end{equation}

\begin{equation}
    \label{eq:mix-ce-simplified}
    L_{ce}^{mix}(X,Y) = 0.5 \cdot L_{ce}(X,Y) + 0.5 \cdot L_{ce}(X,Y_m)
\end{equation}

\begin{equation}
    Y_m = \{y_i\}^N | \{y_j\}^N
\end{equation}


we can derive a similar compact form for supervised contrastive loss under Mixsum setting in Equation \ref{eq:mix-sup-simplified}. The derivation is inspired by the cross entropy loss under Mixsum setting.

\begin{equation}
    \label{eq:mix-sup-simplified}
    L_{sup}^{mix}(X,Y) \approx 0.5 \cdot L_{sup}(X,Y) + 0.5 \cdot L_{sup}(X,Y_m)
\end{equation}

The constraints $1_{y_i=y_j}$ in Equation 7 can be written as $y_i \cdot y_j$, where $y_i$ and $y_j$ are the one hot label vectors. Then in the Mixsum setting, each mixed label $y_i^{mix}$ is obtained by $0.5 \cdot y_i + 0.5 \cdot y_i^m$, where $y_i \in Y$ and $y_i^m \in Y_m$. Thus, by expanding the LHS of Equation \ref{eq:mix-sup-simplified}, we can replace the constraints $1_{y_i^{mix} = y_j^{mix}}$ with $y_i^{mix} \cdot y_j^{mix}$, which is 
\begin{equation}
    \label{eq:sup-mix-1}
    (0.5 \cdot y_i + 0.5 \cdot y_i^m) \cdot (0.5 \cdot y_j + 0.5 \cdot y_j^m)
\end{equation}
Expanding Equation \ref{eq:sup-mix-1}, we can get
\begin{equation}
    \label{eq:mixsum-complex-label}
    0.25(y_i \cdot y_j + y_i \cdot y_j^m + y_i^m \cdot y_j + y_i^m \cdot y_j^m)
\end{equation}
But Equation \ref{eq:mixsum-complex-label} is too complex for computation and also not neat, so we decided to do an approximation--using $y_i \cdot y_j + y_i^m \cdot y_j^m$ to approximate $y_i \cdot y_j^m + y_i^m \cdot y_j$. Then we can get 
\begin{equation}
    \begin{aligned}
    y_i^{mix} \cdot y_j^{mix} & \approx 0.5(y_i \cdot y_j + y_i^m \cdot y_j^m) \\
    & \approx 0.5(1_{y_i=y_j} + 1_{y_i^m=y_j^m})
    \end{aligned}
\end{equation}
Benefit of doing this approximation is that it can reduce the complexity and make final form neat, and we commit that this approximation inevitably will lose some information.

Minimizing Equation \ref{eq:mix-sup-simplified} is sufficient to achieve the goal--pull the representation of Mixsum sample "in between" the representation of class $y_j$ and $y_i$. 

Finally, combining the cross-entropy loss and supervised contrastive loss under the Mixsum setting, we can get the final objective in Equation \ref{total-loss-mix}.
\begin{equation}
    \label{total-loss-mix}
    L^{mix}(X,Y) = \lambda L_{ce}^{mix}(X,Y) + (1-\lambda)L_{sup}^{mix}(X,Y)
\end{equation}

\section{Experiments}
\label{sec:experiment}
\subsection{Datasets}
We use Amazon-5, Yelp-5, AG News and IMDb text classification datasets for benchmarking, and the dataset splits are obtained from \citet{zhang_character-level_2015}.  

In order to demonstrate the effectiveness of the proposed methods under the limited annotation setting, we randomly sample ten subsets using ten different random seeds from each of Amazon-5, Yelp-5, AG-News and IMDb for each experiment, each subset contains 80 training samples and 1000 test samples. The statistics of sampled datasets is shown in Table \ref{tab:data-stat}.
\begin{table}[htbp]
    \centering
    \begin{tabular}{ccccc}
        \toprule
         Dataset & Train set & Test set & \#Class\\
         \midrule
        Amazon (S) & 80 & 1000 & 5 \\
        Yelp (S) & 80 & 1000 & 5 \\
         AG-News (S) & 80 & 1000 & 4 \\
        IMDb (S) & 80 & 1000 & 2 \\
         \toprule
    \end{tabular}
    \caption{Dataset statistics.
    (S) denotes the dataset sampled with small number of train samples.
    }
    \label{tab:data-stat}
\end{table}

\subsection{Experimental Setting}
\label{sec:setting}
For all the experiments, we test the proposed methods using several pretrained transformer models as backbone text-feature encoders including Roberta-base model\cite{DBLP:journals/corr/abs-1907-11692}, and Bert-base model\cite{devlin-etal-2019-bert}. As for the pooling strategy of the backbone encoder, we simply use the feature of [CLS] token as the sentence feature, which is commonly used as the text feature for text classification. Adam optimizer \cite{DBLP:journals/corr/KingmaB14} is used for optimization. The maximum learning rate is set to $1e-5$, and the learning rate is decayed linearly with warm-up steps. The batch size is set to 8. 
We set the trade-off parameter $\lambda$ to $0.9$ for experiment involving $L_{sup}$, since $0.9$ is the optimal trade-off parameter between supervised contrastive loss and cross-entropy loss when using Back-Translation for augmentation according to \citet{DBLP:journals/corr/abs-2011-01403}.

The summarization method we used for creating contrastive samples is PreSumm \cite{liu-lapata-2019-text}, which is available on github\footnote{\url{https://github.com/nlpyang/PreSumm}}, and we also use the Text-Rank algorithm for replacement when junk outputs are generated by PreSumm. It's inevitable for abstractive summarization methods like PreSumm to generate some junk outputs when certain input texts are given, and only a few junk outputs will be generated. Text-Rank is an extractive summarization method, which generates summaries by extracting existing sentences in the texts.

All of our code and datasets are available on the github repository\footnote{\url{https://github.com/ChesterDu/Contrastive_summary}}.

\subsection{Baselines}
\label{sec:baseline}
In order to testify the effectiveness of creating contrastive samples using summarization, we compare the proposed data augmentation strategy with Back-Translation\cite{edunov-etal-2018-understanding}. Back-Translation is a common data augmentation strategy for contrastive learning in NLP\cite{fang_cert_2020}. We first translate the training samples in English to Chinese and then translate back the Chinese texts to English using Google Translate.

We also conduct an ablation experiment under a setting that does not use summarization as contrastive samples. Under this setting, we simply remove the augmented samples in the data batch and only use original samples in the batch. The objective function under this setting only consists of cross-entropy loss and supervised contrastive loss of original samples.

\subsection{Results}
All the experiment results reported are the average results of repeating experiments with ten different random seeds. The experiment settings for producing all the results are introduced in Section \ref{sec:baseline} and \ref{sec:setting}. 
\subsubsection{Comparison to Baseline}
\begin{table}[h]
    \centering
    \begin{tabular}{p{3.5cm}p{1.5cm}p{1cm}p{1cm}}
        \toprule
        Methods  & Bert & Roberta \\
        \midrule
        Amazon(S)  & & \\               
        \hline
        BT  & 31.6 & 28.7 \\
        Sum   & 33.4 & 30.0 \\
        Mixsum & \textbf{34.1} & \textbf{35.2} \\

        \toprule
        Yelp(S) & & \\
        \hline
        BT  & 36.4 & 35.7 \\
        Sum & 38.2 & 39.0 \\
        Mixsum & \textbf{38.9} & \textbf{42.0} \\
        
        \toprule
        AG-News(S) & & \\
        \hline
        BT  & 81.9 & 74.5 \\
        Sum  & 82.3 & 76.2 \\
        Mixsum  & \textbf{83.7} & \textbf{76.5} \\
        
        \toprule
        IMDb(S) & & \\
        \hline
        BT  & 74.5 & 85.6 \\
        Sum  & 75.1 & 87.3 \\
        Mixsum & \textbf{76.6} & \textbf{87.7} \\
        
        \toprule
    \end{tabular}
    \caption{Comparison to Back-Translation baseline.
    BT denotes the setting that using Back-Translation to create contrastive samples.
    Sum denotes the setting that using summarization to create contrastive samples proposed by us.
    Mixsum denotes the setting that using Mixsum for supervised contrastive learning.
    }
    \label{tab:baseline-results}
\end{table}

We have two findings from the experiment results in Table \ref{tab:baseline-results}. First, the proposed contrastive samples generation technique, i.e. summarization, outperforms the Back-Translation method\cite{edunov-etal-2018-understanding} under limited annotation setting on all four datasets. Second, the proposed Mixsum method can further improve the performance of using summarization for contrastive samples generation(Sum).

\subsubsection{Ablation Study}
\label{sec:exp:ablation}
In order to demonstrate the effectiveness of the proposed two methods, we conduct ablation experiments on Amazon(S), Yelp(S), AG-News(S) and IMDb(S) to see the classification accuracy gain of each methods. The results are shown in Table \ref{tab:ablation-results-amazon}, \ref{tab:ablation-results-yelp}, \ref{tab:ablation-results-ag_news} and \ref{tab:ablation-results-imdb}. $L_{ce}$ represents the setting that only use cross entropy loss and without any data augmentation. $L_{ce} + L_{sup}(N)$ represents the setting that do not use summarization as contrastive samples, and only use original samples for supervised contrastive learning. Under this setting, we can simply remove the augmented samples in the data batch and only use original samples in the minibatch. $L_{ce} + L_{sup}(Sum)$ represents the setting that uses summarization to create contrastive samples, which is introduced in Section \ref{sec:methods-sup}. $L_{ce} + L_{sup}(Sum+BT)$ represents the setting that combine summarization and Back-Translation together for contrastive samples generation. $L_{ce}^{mix} + L_{sup}^{mix}$ is the setting that uses Mixsum introduced in Section \ref{sec:methods-mixsum} for supervised contrastive learning.


\begin{table}[h]
    \centering
    \begin{tabular}{p{4cm}p{1.5cm}p{1cm}p{1cm}}
        \toprule
        Methods & Bert & Roberta \\
        \midrule
        $L_{ce}$              & 30.5 & 29.1 \\
        $L_{ce} + L_{sup}(N)$  & 31.5 & 28.0 \\
        $L_{ce} + L_{sup}(Sum)$     & 32.5 & 30.0 \\
        $L_{ce} + L_{sup}(Sum+BT)$     & 29.1 & 25.3 \\
        $L_{ce}^{mix} + L_{sup}^{mix}$  & \textbf{34.1} & \textbf{35.2} \\

        \toprule
    \end{tabular}
    \caption{Ablation  Results on Amazon(S),.
    }
    \label{tab:ablation-results-amazon}
\end{table}

\begin{table}[h]
    \centering
    \begin{tabular}{p{4cm}p{1.5cm}p{1cm}p{1cm}}
        \toprule
        Methods & Bert & Roberta \\
        \midrule
        $L_{ce}$              & 34.1 & 35.9 \\
        $L_{ce} + L_{sup}(N)$ & 34.9 & 36.7 \\
        $L_{ce} + L_{sup}(Sum)$     & 38.2 & 39.0 \\
        $L_{ce} + L_{sup}(Sum+BT)$     & 34.6 & 37.1 \\
        $L_{ce}^{mix} + L_{sup}^{mix}$ & \textbf{38.9} & \textbf{42.0} \\
        
        \toprule
    \end{tabular}
    \caption{Ablation  Results on Yelp(S).
    }
    \label{tab:ablation-results-yelp}
\end{table}

\begin{table}[h]
    \centering
    \begin{tabular}{p{4cm}p{1.5cm}p{1cm}p{1cm}}
        \toprule
        Methods & Bert & Roberta \\
        \midrule
        $L_{ce}$               & 79.9 & 74.2 \\
        $L_{ce} + L_{sup}(N)$  & 80.1 & 70.7 \\
        $L_{ce} + L_{sup}(Sum)$    & 82.3 & 76.2 \\
        $L_{ce} + L_{sup}(Sum+BT)$     & 81.5 & 74.3 \\
        $L_{ce}^{mix} + L_{sup}^{mix}$ & \textbf{83.7} & \textbf{76.5} \\
        
        \toprule
    \end{tabular}
    \caption{Ablation  Results on AG-News(S).
    }
    \label{tab:ablation-results-ag_news}
\end{table}

\begin{table}[h]
    \centering
    \begin{tabular}{p{4cm}p{1.5cm}p{1cm}p{1cm}}
        \toprule
        Methods & Bert & Roberta \\
        \midrule
        $L_{ce}$               & 72.2 & 83.8 \\
        $L_{ce} + L_{sup}(N)$  & 72.9 & 85.6 \\
        $L_{ce} + L_{sup}(Sum)$     & 75.1 & 87.3 \\
        $L_{ce} + L_{sup}(Sum+BT)$     & 71.4 & 86.2 \\
        $L_{ce}^{mix} + L_{sup}^{mix}$  & \textbf{76.6} & \textbf{87.7} \\
        
        \toprule
    \end{tabular}
    \caption{Ablation  Results on IMDb(S).
    }
    \label{tab:ablation-results-imdb}
\end{table}

We have four findings from the Ablation Results.
\begin{itemize}
    \item The proposed summarization method can significantly increase the performance, and the average performance gain is $2.61\%$ across all datasets and models compared to $L_{ce}$ setting.
    \item The proposed Mixsum method can further improve the performance of the classifier. The average performance gain compared to $L_{ce}$ setting is $4.38\%$, and the average performance gain compared to the summarization method is $1.7\%$.
    \item Supervised contrastive learning without any augmented contrastive samples may or may not increase the classifier performance, the average performance gain is $0.0875\%$ across all datasets and models. Sometimes it would even decrease the performance of classifier.
    \item Combining Sum and BT samples together can not outperforms the setting that only use one of them.
\end{itemize}


\subsubsection{Sensitive analysis}
In order to investigate how the number of training examples impacts the performance of the proposed methods, we report the test accuracy on datasets with 800 and 6500 training examples. The trade-off parameter $\lambda$ is set to $0.99$. We only conduct the experiment using the Roberta-base model for convenience since we think that the results obtained from Roberta are representative enough according to Ablation Results. The results is shown in Table \ref{tab:results-800} and \ref{tab:results-6500}.



\begin{table}[h]
    \centering
    \begin{tabular}{p{2.5cm}p{1.5cm}p{1cm}p{1cm}}
        \toprule
        Methods & Amazon(M) & Yelp(M) & AG(M) \\
        \midrule
        $L_{ce}$               & 57.4 & 57.8 & 87.7 \\
        $L_{ce} + L_{sup}$(N)  & 57.6 & 57.4 & 87.4 \\
        $L_{ce} + L_{sup}$     & 56.7 & 58.1 & 87.4 \\

        $L_{ce}^{mix} + L_{sup}^{mix}$  & \textbf{58.1} & \textbf{58.2} & \textbf{88.8} \\
        \toprule
    \end{tabular}
    \caption{Test Accuracy on datasets with 800 training examples.
            (M) denotes the dataset sampled with 800 train samples.
            }
    \label{tab:results-800}
\end{table}

\begin{table}[h]
    \centering
    \begin{tabular}{p{2.8cm}p{1.5cm}p{1cm}p{1cm}}
        \toprule
        Methods & Amazon(L) & Yelp(L) & AG(L) \\
        \midrule
        $L_{ce}$               & 84.8 & \textbf{61.0} & 95.9 \\
        $L_{ce} + L_{sup}(N)$  & 84.6 & 59.8 & 95.9 \\
        $L_{ce} + L_{sup}(Sum)$     & \textbf{84.8} & 60.4 & 95.7 \\
        $L_{ce}^{mix} + L_{sup}^{mix}$  & 84.0 & 60.6 & \textbf{96.3} \\
        \toprule
    \end{tabular}
    \caption{Test Accuracy on datasets with 6500 training examples
            (L) denotes the dataset sampled with 6500 train samples.}
    \label{tab:results-6500}
\end{table}

We observe that when the number of training samples increases, Mixsum still can achieve better performance in those three datasets compared to ablation methods. However, compared to results when the number of training samples is only $80$ in Section \ref{sec:exp:ablation}, we find that performance improvement of the proposed two methods is much smaller. When the number of training samples increases to 6500, the performance of the proposed methods even lower than the ablation setting. Combining results from Section \ref{sec:exp:ablation}, it's reasonable to infer that the proposed methods are beneficial under the limited annotation scenario, but they may not necessary when the number of training samples get larger.

In order to investigate how summarization methods will impact the performance of the proposed methods, we replace the original abstractive summarization method--PreSumm\cite{liu-lapata-2019-text} with extractive summarization method--TextRank. TextRank algorithm will rank the relative importance of the sentences in a text and select the most important sentence as the text summary. We report the test accuracy of using TextRank for text summarization in Table \ref{tab:results-ext_80}.
\begin{table}[h]
    \centering
    \begin{tabular}{p{2.8cm}p{1.5cm}p{1cm}p{1cm}}
        \toprule
        Methods & Amazon(S) & Yelp(S) & AG(S) \\
        \midrule
        $L_{ce}$               & 29.1 & 35.9 & 74.2 \\
        $L_{ce} + L_{sup}(N)$  & 28.0 & 36.7 & 70.7 \\
        $L_{ce} + L_{sup}(Sum)$     & 26.7 & 38.5 & 75.7 \\

        $L_{ce}^{mix} + L_{sup}^{mix}$  & \textbf{29.6} & \textbf{41.2} & \textbf{76.2} \\
        \toprule
    \end{tabular}
    \caption{Text Accuracy on datasets by using extractive summarization}
    \label{tab:results-ext_80}
\end{table}

With this alternative summarization system, the performance of the proposed mix-sum regularization methods is not as good as using PreSumm. We think that it is the limitation of the extractive summarization that leads to the performance drop because extractive summarization can only create summaries from original texts and will bring more information loss compared to abstractive summarization. Besides, the performance of the proposed Mixsum regularization still outperforms other ablation models, which proved the generalization ability of the proposed Mixsum method over different summarization methods.

Furthermore, we also investigated effect of using different texts mixing methods. \citet{sun-etal-2020-mixup} propose to mix the texts by linearly interpolating sentence-level features of texts. The sentence-level features are encoded by a pre-trained transformer model, like BERT and Roberta. We replace our texts mixing methods with the linear interpolation of sentence-level feature as introduced by \citet{sun-etal-2020-mixup}, and keep all other settings same as Mixsum introduced in Section \ref{sec:method} and \ref{sec:experiment}. The results are shown in Table \ref{tab:results-mix_methods}. All the experiment are repeated with $10$ different random seeds.

\begin{table}[h]
    \centering
    \begin{tabular}{p{2.8cm}p{1.5cm}p{1cm}p{1cm}}
        \toprule
        Methods & Amazon(S) & Yelp(S) & AG(S) \\
        \midrule
        Sum           & 30.0 & 39.0 & 76.2 \\
        Mixsum(Ours)  & 35.2 & 42.0 & 76.5 \\
        Mixsum(LISF)  & 32.5 & 41.1 & 77.3 \\
        \toprule
    \end{tabular}
    \caption{Comparison of using linear interpolation of sentence-level feature(LISF) as texts mixing methods with concatenation of summary texts(Ours).}
    \label{tab:results-mix_methods}
\end{table}

We observe that replacing our texts mixing methods with LISF still can achieve similar results and outperforms the Sum setting. Thus, we believe that other different sentence mixing methods can also be adopted in Mixsum framework.

\section{Conclusion}
\label{sec:conclusion}
We proposed a novel data augmentation technique for constructing contrastive samples in supervised contrastive learning--summarization. Besides, we also proposed a Mixsum method based on using summarization to construct the contrastive samples. We demonstrate the effectiveness of the proposed two new techniques on text classification task under the limited annotation setting. The experiment results on four datasets show that Mixsum and using summarization as contrastive samples can improve the performance of text classification under the limited annotations setting. Besides, We show that the proposed Mixsum methods can be generalized to different summarization methods and text mixing methods.

Our work also opens up several possibilities for future work, since using summarization to construct contrastive samples has shown the effectiveness in supervised contrastive learning. We may investigate whether using summarization as data augmentation can improve unsupervised text classification \cite{wu2018word}, and the robustness and performance of other NLP applications like question answering, commonsense reasoning and semantic code retrieval\cite{ling2021deep, ling2021multilevel}.





\bibliography{anthology,custom}
\bibliographystyle{acl_natbib}

\clearpage
\newpage
\onecolumn
\appendix
\section{Pseudo-code}
Pseudo-code of using summaries for supervised contrastive learning, mentioned in Section \ref{sec:methods-sup}.
\begin{algorithm*}[!htp]
\label{alg:scl}
    \caption{Pseudo-code of Sum}
    \begin{algorithmic}
\Require Initialize the backbon encoder $f$, classification head $proj$ and mlp head $mlp$. \\Trade off parameter $\lambda$.
\For {sampled minibatch $\{x_k\}_{k=1}^{N}$, $\{y_k\}_{k=1}^{N}$}
    \For {$k\in \{1,...N\}$} \
        \State $\hat{x}_{2k} = Summ(x_k)$     \hfill{\textcolor{gray}{// use summarization for augmentation}}    
        \State $z_{2k} = f(\hat{x}_{2k-1})$  \hfill{\textcolor{gray}{// get backbone representation}}
        \State $s_{2k} = proj(z_{2k-1})$  \hfill{\textcolor{gray}{// project the representation to prediction score}}
        \State $g_{2k} = mlp(z_{2k-1})$  \hfill{\textcolor{gray}{// Apply MLP head to get feature representation of summary}}
        \State $g_{2k} = Norm(g_{2k})$ \hfill{\textcolor{gray}{// Normalize the feature vector}}
        \\
        \State $\hat{x}_{k} = x_k$ \hfill{\textcolor{gray}{// original texts}}
        \State $z_{k} = f(\hat{x}_{k})$ \hfill{\textcolor{gray}{// get backbone representation}}
        \State $s_{k} = proj(z_{k})$  \hfill{\textcolor{gray}{// project the representation to prediction score}}
        \State $g_{k} = mlp(z_{k})$ \hfill{\textcolor{gray}{// Apply MLP head to get feature representation of summary}}
        \State $g_{k} = Norm(g_{k})$ \hfill{\textcolor{gray}{// Normalize the feature vector}}
        \\
        \State $\hat{y}_{2k} = y_k$ \hfill{\textcolor{gray}{//the label of summary is same as original text}}
        \State $\hat{y}_{k} = y_k$
    \EndFor
    \State $l_{ce} = CrossEnropy(\{s_k\}_{k=1}^{2N},\{\hat{y}_k\}_{k=1}^{2N})$ \hfill{\textcolor{gray}{// cross entropy loss of augmented batch}}
    \State $l_{sup} = SupConLoss(\{g_k\}_{k=1}^{2N},\{\hat{y}_k\}_{k=1}^{2N}))$ \hfill{\textcolor{gray}{// contrastive loss of augmented batch}}
    \State $L = \lambda l_{ce} + (1-\lambda)l_{sup}$ \hfill{\textcolor{gray}{//compute total loss}}
    \\
    \State Compute $\nabla _ {\theta _{f}} L$, $\nabla _ {\theta _{proj}} L$, $\nabla _ {\theta _{mlp}} L$
    \State Update $f,proj,mlp$ to optimize $L$
    \\
\EndFor
    \end{algorithmic}
    
\end{algorithm*}

    
    
    

\clearpage
\newpage
Pseudo-code of using Mixsum for supervised contrastive learning, mentioned in Section \ref{sec:methods-mixsum}. 
\begin{algorithm*}[!htp]
\label{alg:mixsum}
    \caption{Pseudo-code of Mixsum} 
    \begin{algorithmic}
\For {sampled minibatch $\{x_k\}_{k=1}^{N}$, $\{y_k\}_{k=1}^{N}$}
    \State $perm\_index = shuffle(\{1...N\})$
    \For {$k\in \{1,...N\}$}
        \State $j = perm\_index[k]$ \hfill{\textcolor{gray}{// permutation index}}
        \\
        \State $\hat{x}_{2k} = Summ(x_k) + Summ(x_j)$ \hfill{\textcolor{gray}{// use mix-sum for augmentation}}
        \State $z_{2k} = f(\hat{x}_{2k})$ \hfill{\textcolor{gray}{// get backbone representation}}
        \State $s_{2k} = proj(z_{2k})$ \hfill{\textcolor{gray}{// project the representation to prediction score}}
        \State $g_{2k} = mlp(z_{2k})$ \hfill{\textcolor{gray}{//Apply MLP head to get feature vector}}
        \State $g_{2k} = Norm(g_{2k})$ \hfill{\textcolor{gray}{//Normalize the feature vector}}
        \\
        \State $\hat{x}_{k} = x_k$ \hfill{\textcolor{gray}{//original texts}}
        \State $z_{k} = f(\hat{x}_{k})$ \hfill{\textcolor{gray}{// get backbone representation}}
        \State $s_{k} = proj(z_{k})$ \hfill{\textcolor{gray}{// project the representation to prediction score}}
        \State $g_{k} = mlp(z_{k})$ \hfill{\textcolor{gray}{//Apply MLP head to get feature vector}}
        \State $g_{k} = Norm(g_{2k})$ \hfill{\textcolor{gray}{//Normalize the feature vector}}
        \\
        \State $\hat{y}_{k} = y_k$ \hfill{\textcolor{gray}{// mix the label}}
        \State $\hat{y}_{2k} = y_k$
        \State $\tilde{y}_{k} = y_k$
        \State $\tilde{y}_{2k} = y_j$
    \EndFor
    \State $l_{ce} = CrossEnropy(\{s_k\}_{k=1}^{2N},\{\hat{y}_k\}_{k=1}^{2N})/2$
    \State $l_{ce} += CrossEnropy(\{s_k\}_{k=1}^{2N},\{\tilde{y}_k\}_{k=1}^{2N})/2$
    \State $l_{scl} = SupConLoss(\{g_k\}_{k=1}^{2N},\{\hat{y}_k\}_{k=1}^{2N}))/2$
    \State $l_{scl} += SupConLoss(\{g_k\}_{k=1}^{2N},\{\tilde{y}_k\}_{k=1}^{2N}))/2$
    \State $L = \lambda l_{ce} + (1-\lambda) l_{scl}$
    \\
    \State Compute $\nabla _ {\theta _{f}} L$, $\nabla _ {\theta _{proj}} L$, $\nabla _ {\theta _{mlp}} L$
    \State Update $f,proj,mlp$ to optimize $L$
    \\
\EndFor
    \end{algorithmic}
    
\end{algorithm*}

    
    
    



\end{document}